\documentclass[letterpaper]{article}
\usepackage{aaai18}
\usepackage{times}
\usepackage{helvet}
\usepackage{courier}
\usepackage{url}
\usepackage{siunitx}

\usepackage{graphicx}
\usepackage{subcaption}
\usepackage{amsmath}
\usepackage{amssymb}
\usepackage{amsthm}
\usepackage{algorithm}
\usepackage{algpseudocode}
\usepackage{optidef}
\usepackage{mathtools}
\usepackage{multirow}
\usepackage{array}

\newcommand{\citet}[1]
{\citeauthor{#1}~\shortcite{#1}}
\newcommand{\citep}{\cite}

\title{Challenges of Context and Time in Reinforcement Learning:\\ Introducing Space Fortress as a Benchmark}
\author{}
\author{Akshat Agarwal\\
Robotics Institute\\
Carnegie Mellon University\\
\texttt{aa7@cmu.edu}\\
\And
Ryan Hope\\
Department of Psychology\\
Carnegie Mellon University\\
\texttt{rmh3093@gmail.com}\\
\And
Katia Sycara\\
Robotics Institute\\
Carnegie Mellon University\\
\texttt{katia@cs.cmu.edu}\\
}

\begin{document}
\maketitle
\begin{abstract}
Research in deep reinforcement learning (RL) has coalesced around improving performance on benchmarks like the Arcade Learning Environment. However, these benchmarks conspicuously miss important characteristics like abrupt context-dependent shifts in strategy and temporal sensitivity that are often present in real-world domains. As a result, RL research has not focused on these challenges, resulting in algorithms which do not understand critical changes in context, and have little notion of real world time. To tackle this issue, this paper introduces the game of Space Fortress as a RL benchmark which incorporates these characteristics. We show that existing state-of-the-art RL algorithms are unable to learn to play the Space Fortress game. We then confirm that this poor performance is due to the RL algorithms'  context insensitivity and reward sparsity. We also identify independent axes along which to vary context and temporal sensitivity, allowing Space Fortress to be used as a testbed for understanding both characteristics in combination and also in isolation. We release Space Fortress as an open-source Gym environment.
\end{abstract}

\section{Introduction}

Recent advances in computer vision \cite{krizhevsky2012imagenet} and natural language processing \cite{sutskever2014sequence} can be attributed to the advent of deep learning and the presence of robust benchmarks to quantitatively measure progress, such as the ImageNet challenge \cite{russakovsky2015imagenet}. In the last few years, neural network-based function approximation has also proven successful in reinforcement learning, with AI agents now able to perform at superhuman levels in games like Go \cite{silver2016mastering} and the Atari \cite{mnih2015human} suite. Once again, research in Deep RL has been steered by the establishment of benchmarks like the Arcade Learning Environment \cite{bellemare2013arcade}, along with the OpenAI Gym interface \cite{brockman2016openai}, which has been widely adopted by the research community. 

These benchmarks are conspicuously missing 2 challenging characteristics: (a) abrupt context-dependent switching of strategy and (b) temporal sensitivity. For agents to operate in the real world, they need to be able to switch behaviors very abruptly, which necessitates (i) learning to identify critical points where behavior needs to change, and (ii) learning the different behaviors required in each context. Agents also need to have an understanding of time as an independent variable, along with the ability to adapt their behavior accordingly. While having no understanding of time as something that's always ticking might work for simulated or static real-world environments, it is not acceptable for real-world dynamic environments with moving entities and where decisions might have to be adaptively taken very quickly or very slowly, depending on the context.
Since existing benchmarks do not focus on these properties, reinforcement learning research has not tackled these problems yet.

In this paper, we introduce a challenging RL environment based on Space Fortress (SF) \cite{mane1989space}, an arcade-style game which was developed by psychologists in the 80s to study human skill acquisition, and is still used quite frequently \cite{towne2016understanding,destefano2016should}. The objective of the game is to fly a ship and destroy a fortress by firing missiles at it. The ship has to respect a minimum time difference between successive shots, while building up the fortress' vulnerability, and once the fortress becomes vulnerable, destroy it with a rapid double shot. As a RL testbed, Space Fortress possesses both the characteristics discussed above: context-dependent strategy change (change in required firing rate after the fortress becomes vulnerable) and time sensitivity (firing rate requirements independent of the agent's decision speed i.e., the frame rate). It also has a sparse reward structure, and, as we show, is not solved by any state-of-the-art RL algorithms such as Rainbow \cite{hessel2017rainbow}, Proximal Policy Optimization (PPO) \cite{schulman2017proximal} and Advantage Actor-Critic (A2C) \cite{mnih2016asynchronous}. 

While being an interesting and relevant challenge for reinforcement learning, the rich background on human skill acquisition research based on Space Fortress also makes it an attractive tool to study human-AI collaboration in a dynamic environment, compare skill acquisition techniques of humans vs artificial agents, and work on few-shot learning by leveraging lessons from cognitive architectures like ACT-R \cite{anderson2009can} which have previously learned the game with extremely high sample efficiency, albeit using handcrafted features and extensive domain knowledge.

We make the following contributions. First, we present a new RL testbed that requires the agent to switch strategies abruptly based on context, and develop a conceptualization of time independent of its speed of decision making, and demonstrate empirically that performance on par with humans is beyond the capability of current state-of-the-art RL algorithms, even after relaxing the reward sparsity through shaping \cite{ng1999policy}. We identify the aspects of the game which can be varied to control both temporal and context sensitivity, allowing research on either \textit{in isolation}. Finally, we demonstrate that \textit{after introducing modifications} to ease identification of critical contexts, the PPO algorithm learns to play the game well enough to outperform humans, verifying that context insensitivity is the primary driver behind the poor performance of RL algorithms. We also present robust human benchmark results for Space Fortress, allowing future researchers to place new experimental results in context. We open-source\footnote{\url{https://github.com/agakshat/spacefortress}}
the OpenAI Gym environment for Space Fortress as well as all the code used to run our experiments, to promote research in temporal and context-sensitive reinforcement learning algorithms.

\section{Related Work}
\label{sec:relatedwork}
The Arcade Learning Environment (ALE) \cite{bellemare2013arcade} poses the challenge of building AI agents with competency across dozens of Atari 2600 games, like Space Invaders, Asteroids, Bowling and Enduro. Following the development of Deep Q Networks \cite{mnih2015human}, a lot of research in the RL community has focused on improving performance in one or more of the games with improvements like massive parallelization, sample efficiency \cite{wang2015dueling,schaul2015prioritized}, better exploration \cite{fortunato2017noisy,plappert2017parameter}, reward sparsity \cite{pathak2017curiosity,andrychowicz2017hindsight} and long-term strategies \cite{bacon2017option,kulkarni2016hierarchical}. In continuous control tasks on the MuJoCo testbed \cite{todorov2012mujoco}, on-policy actor critic methods \cite{schulman2017proximal,mnih2016asynchronous} have shown promise. \citet{bellemare2017distributional} estimated a probability distribution over the Q-value of a state (instead of just the mean of the Q-value), with greatly improved results. Rainbow \cite{hessel2017rainbow} combined a lot of orthogonal improvements in DQNs to achieve state of the art results. However, we show below that these algorithms fail to learn anything on Space Fortress. 

Games like Ms. Pacman and Seaquest in the ALE have previously required some context or temporal sensitivity, but these characteristics can't be controlled or varied, and form a minor part of the overall game. As a RL testbed, Space Fortress relies heavily on both context and temporal sensitivity, as we show in Section \ref{sec:experiments}, and both characteristics can be controlled directly to enable their study in isolation.

There has also been a fair amount of prior work on reinforcement learning with sparse rewards.  \citet{pathak2017curiosity} use curiosity as an intrinsic reward signal to efficiently direct exploration. State visitation counts have also been investigated for exploration \cite{bellemare2016unifying}, and \citet{osband2016deep} train multiple value functions and make use of bootstrapping and Thompson sampling for exploration. These works focus on learning with sparse rewards through better exploration of the state space, which does not help with Space Fortress where exploration is required in time and in latent contexts.

\citet{zambrano2015continuous} trained agents to deal with actions that take a finite amount of time through neural reinforcement learning in grid worlds, which still did not require a conceptualization of time independent of the internal speed of decision making, hence differing from the proposed work. Finally, \citet{van2017higher,van2017towards} previously used A3C on a simple control task abstracted from Space Fortress, with no fortress destruction required. Crucially, this task removed the interesting characteristics of Space Fortress, namely contextual and temporal sensitivity, as well as reward sparsity. We release an implementation of the game as an OpenAI Gym environment to promote research, conduct an ablation study to ascertain the roles of context and temporal sensitivity and reward sparsity in poor performance, and then present results showing existing RL algorithms outperforming humans after we control for the above factors.

\section{The Space Fortress RL Environment}
\label{sec:sf}
\begin{figure*}[t]
    \centering
    \begin{subfigure}{0.3\textwidth}
        \centering
        \includegraphics[width=\textwidth]{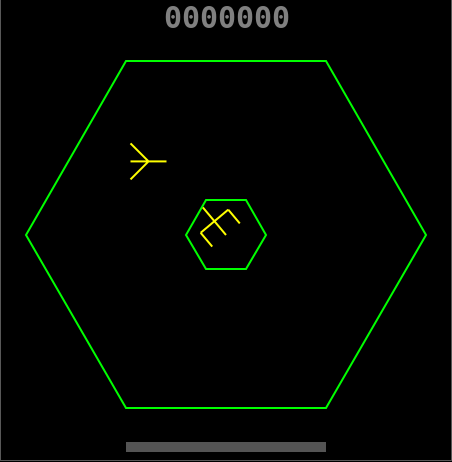}
        \caption{}
    \end{subfigure}
    ~
    \begin{subfigure}{0.3\textwidth}
        \centering
        \includegraphics[width=\textwidth]{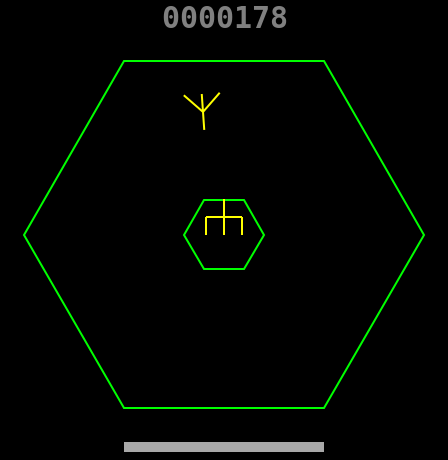}
        \caption{}
    \end{subfigure}
    ~
    \begin{subfigure}{0.3\textwidth}
        \centering
        \includegraphics[width=\textwidth]{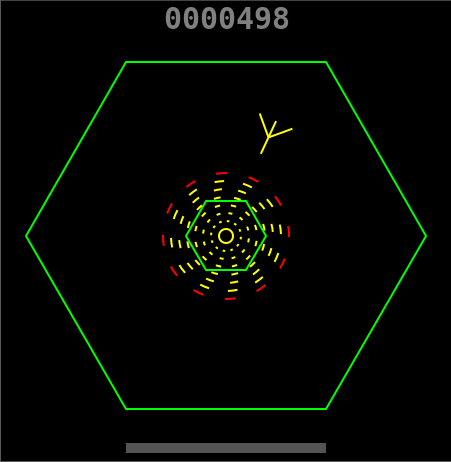}
        \caption{}
    \end{subfigure}
    \caption{Game Screens in \textbf{Space Fortress}. The ship has to fly between the two hexagons, while the fortress can only change its orientation at a fixed position. The game score is displayed at the top, and the fortress' vulnerability is displayed as a bar which fills up on each shot. (a) The bar is empty, indicating that the fortress' vulnerability is 0, (b) The bar is now full, indicating that vulnerability is equal to 10 and a rapid double shot will now destroy the fortress and (c) The fortress has been destroyed. This is followed by a reset of the fortress and continuation of the game till end of episode (3 minutes game time)}
    \label{fig:sf}
\end{figure*}

We now describe the Space Fortress game, discuss its utility as a testbed for reinforcement learning, and present results from humans learning to play the game, intended as a baseline. The game environment can be seen in Fig. \ref{fig:sf}.  

\subsection{Game Description}
\label{sec:gamedescription}
The player/AI agent controls a ship, which has to fly around in a frictionless arena, firing missiles to destroy a fortress located centrally within the arena. Hitting the walls on either sides or being hit by shells fired by the fortress results in immediate ship death, which incurs a penalty on the agent. Destroying the fortress, however, requires a \textit{context-aware strategy}. Each missile that successfully hits the fortress increases its vulnerability $v$ by one. When $v<10$, the fortress is 'not vulnerable', and the ship must fire its missiles spaced \textit{more than 250ms} apart. Firing faster than this while $v<10$ leads to the fortress vulnerability getting reset back to zero. This is obviously undesirable and the agent must learn to shoot slowly. However, once $v = 10$, the fortress becomes vulnerable, and a rapid double fire (2 shots spaced \textit{less than 250ms} apart) is required to destroy the fortress. We refer to this 250ms time specification as the ``critical time interval''.
It is important to note that once $v = 10$, shooting further at the fortress at a rate less than 4Hz will lead to no change in vulnerability. Hence the firing \textit{strategy completely reverses} at the point when vulnerability reaches 10, and the agent must learn to identify this critical point to perform well. 
Since the game is simply reset (without ending the episode) when the fortress is destroyed, it is crucial that the agent also recognize this second critical point of fortress destruction, and switch back its firing rate to continue playing well.
This major dependence on contextual and temporal sensitivity is unique to Space Fortress among RL benchmarks.

A single game lasts for 3 minutes. The game does not end in the event of either a fortress or ship destruction, and points are scored by destroying the fortress as many times as possible in those 3 minutes while avoiding getting shot down by the fortress or colliding with the arena. When a fortress is destroyed, its vulnerability resets to zero, and the game continues. When the ship is destroyed, it respawns at a random position and orientation, but the fortress' vulnerability is preserved.

\subsection{Game Versions}
Space Fortress requires the agent to master advanced controls in a frictionless environment, orienting and firing missiles at the fortress while avoiding shells and not colliding with the walls. Since current RL algorithms proved unable to solve the game in its entirety (see experiments in Section \ref{sec:defaultrewards}), we introduced another version of the game to reduce navigation complexity by having the ship automatically pointed at the fortress. Throughout the rest of the paper, the simpler version is called `Autoturn', while the original game is referred to as `Youturn'.

\subsection{Human Evaluations}
\label{sec:humanevaluations}
The human player results were collected by the authors in the context of a study on human skill acquisition \cite{anderson_betts_bothell_hope_lebiere_2018}.
117 people were asked to play 20 games of Space Fortress, with 52 playing Autoturn and 65 playing Youturn. They were all \textit{given instructions about the rules of the game beforehand}, and told about the change in firing rate required when the fortress vulnerability reaches 10. Considering that humans would require some turns to learn to play the game, we report the following results in Table \ref{tab:humanresults}: (1) Best performance of any subject in any game, (2) Average performance of all subjects in the last 5 games, considering the first 15 as a learning phase, (3) Average performance of all subjects in the last 10 games, considering the first 10 as a learning phase, (4) Average performance of all subjects in the last 15 games, considering the first 5 as a learning phase and (5) Average performance of all subjects in all 20 games. The scores shown to the humans (and reported in Table \ref{tab:humanresults}) were as follows: +100 for fortress destruction, -100 for ship death and -2 for each missile shot to penalize excessive firing. 

\begin{table*}
    \centering
    \begin{tabular}{|c|c|c|c|c|c|c|c|}
        \hline
        \textbf{Game} & \textbf{N} & \textbf{Metric} & \textbf{Best} & \textbf{Last 5} & \textbf{Last 10} & \textbf{Last 15} & \textbf{All}\\
        \hline
        \multirow{2}{*}{Autoturn} & \multirow{2}{*}{52} & Score & 3000 & 1989 & 1978 & 1940 & 1810 \\
         &  & FortressDeath & 40 & 30.311 & 30.044 & 29.591 & 28.181\\
         \hline
        \multirow{2}{*}{Youturn} & \multirow{2}{*}{65} & Score & 2314 & 216 & 153 & 43 & -169 \\
         & & FortressDeath & 32 & 14.36 & 13.704 & 12.882 & 11.4\\
        \hline
    \end{tabular}
    \caption{Aggregated results for 102 humans playing Space Fortress. After being provided with instructions about the rules beforehand, each player played the game for 1 hour, or 20 games. Allowing for a few practice games, we report the average scores on the Last `$K$' games ($K \in \{5,10,15\}$), as well as the best individual score.}
    \label{tab:humanresults}
\end{table*}

\begin{table*}
    \centering
    \begin{tabular}{|c|c|c|>{\centering}m{1.3cm}|>{\centering}m{1.3cm}|m{1.2cm}|}
        \hline
        \textbf{S.No.} & \textbf{Algorithm } & \textbf{Game } & \textbf{Avg. Score } & \textbf{Best Score} & \textbf{Fortress Death} \\
        \hline
        1 & A2C & Autoturn & -2685 & -2242 & 0 \\
        2 & A2C & Youturn & -5859 & --5604 & 0 \\
        \hline
        3 & PPO & Autoturn & -2502 & -2178 & 0 \\
        4 & PPO & Youturn & -5269 & -4698 & 0 \\
        \hline
        3 & Rainbow & Autoturn & -8327 & -8264 & 0 \\
        4 & Rainbow & Youturn & -9378 & -9245 & 0 \\
        \hline
    \end{tabular}
    \caption{Average game scores for RL agents, trained \textbf{with default (sparse) rewards}, for 45M steps}
    \label{tab:originalgameresults}
\end{table*}

\subsection{RL Setup}
\label{sec:rlsetup}
We now describe the exact game setup used for reinforcement learning on Space Fortress.
\begin{itemize}
    \item Observations: The observations are in the form of pixel-level grayscale 84x84 size renderings of the game screen (similar to Fig. \ref{fig:sf}. Important information such as the time lapsed since the last shot is not a part of this observation, making the task partially observed.
    We provide the agent with a stack of the last 4 observations as input at each time step, allowing it to infer direction of movement of the ship and fortress using the difference between successive frames. 
    \item Actions: The agent chooses from 5 actions: (i) No Operation, (ii) Fire (a missile), (iii) Thrust Forward (in the direction of current orientation), (iv) Thrust Right (rotate right without changing position) and (v) Thrust Left (rotate left without changing position). The game operates at a default frame rate of 30 FPS and there is no action repeat, which means an action is chosen every 33ms. Note that the Autoturn version only has 3 actions (since no turning is required).
    \item Rewards: In line with \citet{mnih2015human}, we found that learning was more stable when using clipped rewards. The fortress and ship destruction rewards were clipped to +1 and -1, respectively, and the missile penalty reduced to -0.05. Note that the results used for evaluation and reporting were not clipped, in order to follow the same scheme as described in Section \ref{sec:humanevaluations}.
\end{itemize}

\begin{table*}
    \centering
    \begin{tabular}{|c|c|c|>{\centering}m{1.3cm}|>{\centering}m{1.3cm}|m{1.2cm}|}
        \hline
        \textbf{S.No.} & \textbf{Algorithm } & \textbf{Game } & \textbf{Avg. Score } & \textbf{Best Score} & \textbf{Fortress Death} \\
        \hline
        1 & A2C & Autoturn & -4116 & -2100 & 0 \\
        2 & A2C & Youturn & -4781 & -3890 & 1.3 \\
        \hline
        3 & PPO & Autoturn & -1294 & -1108 & 1 \\
        4 & PPO & Youturn & -1435 & -1206 & 0.94 \\
        \hline
        3 & Rainbow & Autoturn & -6161 & -5960 & 0 \\
        4 & Rainbow & Youturn & -4894 & -4577 & 0 \\
        \hline
    \end{tabular}
    \caption{Average game scores for RL agents, trained \textbf{with dense rewards}, for 45M steps.}
    \label{tab:denseresults}
\end{table*}

\begin{table*}[t]
    \centering
    \begin{tabular}{|c|c|c|c|>{\centering}m{1.3cm}|>{\centering}m{1.3cm}|m{1.2cm}|}
        \hline
        \textbf{S.No.} & \textbf{Algorithm } & \textbf{Architecture} & \textbf{Game } & \textbf{Avg. Score} & \textbf{Best Score} & \textbf{Fortress Death} \\
        \hline
        1 & A2C & SF-GRU & Autoturn & -1641 & -718 & 3 \\
        2 & A2C & SF-GRU & Youturn & -2444 & -1700 & 11 \\
        \hline
        3 & PPO & SF-FF & Autoturn & 2337 & 2818 & 41 \\
        4 & PPO & SF-FF & Youturn & 2235 & 2880 & 40 \\
        \hline
        5 & PPO & SF-GRU & Autoturn & 2510 & 2870 & 43 \\
        6 & PPO & SF-GRU & Youturn & 2356 & 2932 & 41 \\
        \hline
        7 & Rainbow & -- & Autoturn & -2973 & -2330 & 1.2\\
        8 & Rainbow & -- & Youturn & -4112 & -3934 & 0.0\\
        \hline
    \end{tabular}
    \caption{Average game scores for RL agents, trained \textbf{after making context identification easier}, for 45M steps}
    \label{tab:easyresults}
\end{table*}

\section{Experiments and Results}
\label{sec:experiments}
In this section, we experimentally show that (a) no state-of-the-art reinforcement learning algorithm (Rainbow \cite{hessel2017rainbow}, A2C \cite{mnih2016asynchronous} and PPO \cite{schulman2017proximal}) can learn to play Space Fortress, (b) removing reward sparsity does not improve the performance and (c) making context identification easier through specific alterations in the reward structure allow PPO to achieve superhuman performance. We also discuss temporal sensitivity by examining effectiveness of transfer of learning across different settings of the game's critical time interval.

\subsection{Network Architecture}
For Rainbow, the Q-network architecture was identical to that in \citet{hessel2017rainbow}.
For PPO and A2C, we experiment with two policy network architectures:
\begin{itemize}
    \item \textbf{SF-GRU: }
    The agent's policy network takes the 1x84x84 environment observations as input, and outputs (a) a probability distribution over the actions, and (b) a value function estimate of the expected return. The input goes through two convolutional layers with 16, 32 filters of size 8,4 and stride 4,2 respectively, and ReLU activation. The output is flattened and passed through a linear layer with a ReLU non-linearity to get an output vector of size 256. This is then passed through a unidirectional Gated Recurrent Unit (GRU) cell \cite{cho2014learning} with a tanh non-linearity giving an output of size 256. Finally, this vector is passed as input to two linear layers that output the probability distribution over actions (using a softmax activation) and the value estimate of the expected return. 
    \item \textbf{SF-FF: }
    Same as above, but with a fully connected layer of size 256 with ReLU non-linearity instead of the recurrent GRU cell.
\end{itemize}
For all experiments, we ran 16 processes collecting game experience in parallel, with discount factor $\gamma = 0.99$ and Generalized Advantage Estimation (GAE) \cite{schulman2015high} parameter $\lambda = 0.95$. PPO used value loss coefficient $c_{1} = 0.5$, entropy regularization coefficient $c_{2} = 0.05$ and learning rate $\num{1e-3}$, while A2C used $c_{1} = 0.5, c_{2} = 0.01$ and learning rate $\num{5e-4}$. Both A2C and PPO used $n = 1024$-step returns. These hyperparameters were found after extensive tuning. We also clipped the gradients of all the network parameters to 0.5, to prevent catastrophic updates from outlying samples of the expected gradient value. Since the PPO algorithm is more stable, we updated the policy 4 times every epoch - while A2C made only 1 update every epoch.

\begin{figure*}[t]
    \centering
    \begin{subfigure}{0.48\textwidth}
        \centering
        \includegraphics[width=\textwidth]{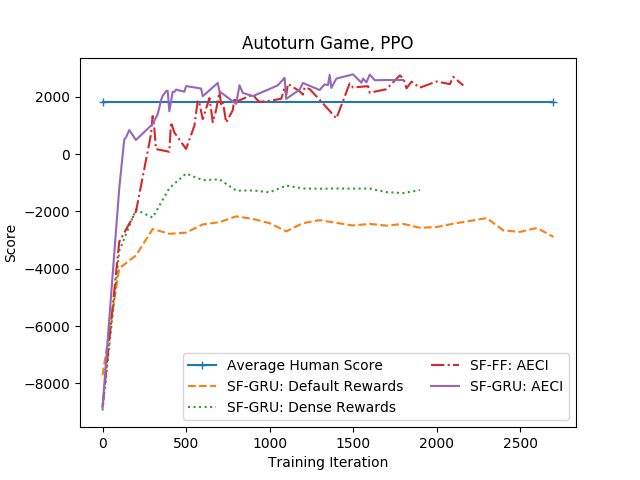}
        \caption{Autoturn}
        \label{fig:ppo_autoturn}
    \end{subfigure}
    ~
    \begin{subfigure}{0.48\textwidth}
        \centering
        \includegraphics[width=\textwidth]{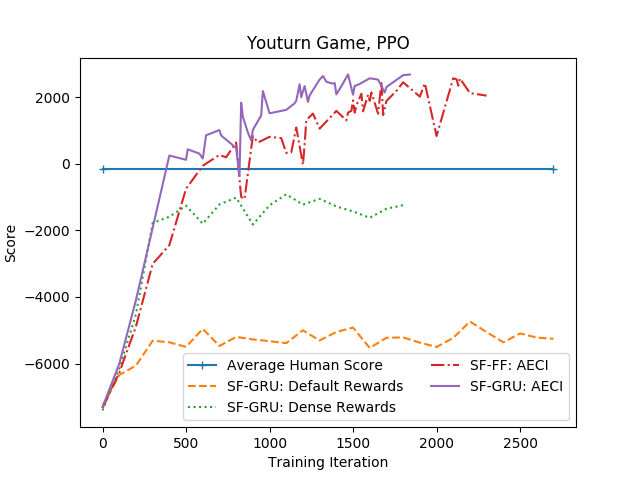}
        \caption{Youturn}
        \label{fig:ppo_youturn}
    \end{subfigure}
    \caption{Learning Curves for PPO on Space Fortress, for different reward structures and architectures. 'Average Human Score' refers to the average score over all 20 games, provided as a point of comparison. 'Default Rewards' discussed in Section \ref{sec:defaultrewards}, 'Dense Rewards' in Section \ref{sec:denserewards} and 'AECI' (After Easing Context Identification) in Section \ref{sec:easingcontextid}. Both SF-FF and SF-GRU architectures are able to achieve superhuman performance \textit{after making context identification easier}. The agent's performance is very poor with both default (sparse) rewards and dense rewards.}
    \label{fig:ppo_curves}
\end{figure*}

\begin{figure*}
    \centering
    \begin{subfigure}{0.32\textwidth}
        \centering
        \includegraphics[width=\textwidth]{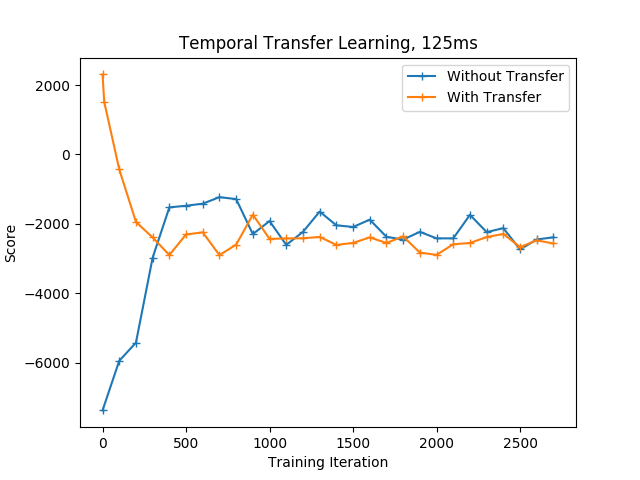}
        \caption{125ms}
        \label{fig:temporal_125}
    \end{subfigure}
    ~
    \begin{subfigure}{0.32\textwidth}
        \centering
        \includegraphics[width=\textwidth]{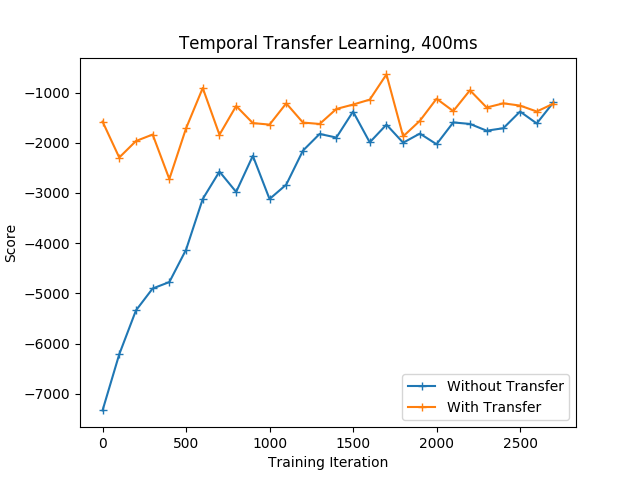}
        \caption{400ms}
        \label{fig:temporal_400}
    \end{subfigure}
    ~
    \begin{subfigure}{0.32\textwidth}
        \centering
        \includegraphics[width=\textwidth]{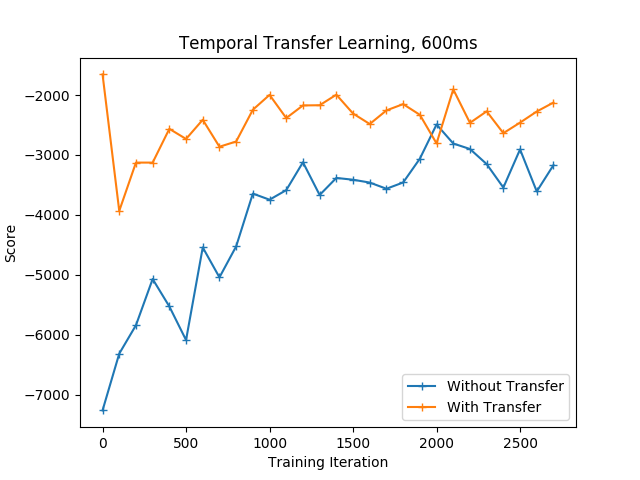}
        \caption{600ms}
        \label{fig:temporal_600}
    \end{subfigure}
    \caption{Checking for positive transfer of learning while changing the critical time interval in Space Fortress. In each figure, we changed the time interval to a different value and verified whether transferring learned weights from the agent trained on 250ms as the critical interval led to any speedup in the learning process, and an improvement in the final performance.}
    \label{fig:temporal}
\end{figure*}

\subsection{With Default (Sparse) Rewards}
\label{sec:defaultrewards}
With the default sparse reward structure which rewards fortress destruction and penalizes ship destruction and missile firing, no algorithm is able to learn to destroy the fortress. A visual inspection of the game play revealed that the PPO and A2C agents (with both architectures) just learned to stop firing, since that leads to an immediate penalty. The Rainbow agent did not learn anything. Table \ref{tab:originalgameresults} presents the aggregated results for PPO, A2C and Rainbow on both versions of the game. The `Fortress Death' column in Table \ref{tab:originalgameresults} indicates the number of times the agent was able to destroy the fortress per game, on average.

\subsection{With Dense Rewards}
\label{sec:denserewards}
Considering the results in Section \ref{sec:defaultrewards} and to understand how reward sparsity is impacting performance, we introduce an additional reward of +1 each time the fortress is hit by a missile, and a penalty of -1 if the fortress' vulnerability gets reset due to a faster firing speed than the context demands. This makes the reward density comparable to Atari games, on which PPO, A2C and Rainbow have all been shown to perform well. 
Their performance on Space Fortress with dense rewards can be seen in Table \ref{tab:denseresults}, where the scores for PPO and Rainbow have improved. From watching a video of the trained agent playing the game, we observed that the improvement stemmed from having learned to avoid ship death and to fire at the fortress, albeit without knowledge of the critical time interval and context-dependent strategy shifts, resulting in an inability to destroy the fortress with any consistency.

Hence, Space Fortress clearly presents a challenge to the state of the art in reinforcement learning, and is a useful and relevant benchmark for further research.

We now move on to studying the impact of context insensitivity of RL algorithms on the task performance (Section \ref{sec:easingcontextid}), and show that by making the identification of critical contexts easier with 2 simple modifications to the reward, PPO learns to play the game very well - outperforming humans comfortably. This clearly indicates that it is context insensitivity and the inability to identify critical points which is hampering performance, further making the case that the Space Fortress game is a useful benchmark for studying context sensitivity.

\subsection{After Making Context Identification Easier}
\label{sec:easingcontextid}
As discussed in Section \ref{sec:gamedescription}, there are 2 critical points which the agent has to learn to identify and switch strategies at. The first is when the fortress becomes vulnerable, i.e. $v = 10$ and the agent has to switch from firing more than 250ms apart to a rapid double shot fired less than 250ms apart. The second is when the fortress is destroyed, and the agent has to switch back to its slow firing speed. To confirm our intuition that it is indeed the algorithms' inability to identify these critical points and accordingly adapt its firing strategy which lead to poor performance, we introduce 2 changes to the reward structure (with respect to the dense reward from Section \ref{sec:denserewards}) which make it trivial for the agent to identify the critical points where context changes:
\begin{itemize}
    \item Instead of rewarding fortress hits (as in Section \ref{sec:denserewards}), we switch to rewarding fortress vulnerability change, by giving a reward of +1 for unit increase in vulnerability, and giving a penalty of -1 for decrease in vulnerability. This has the effect of rewarding fortress hits only until the fortress' vulnerability is building up to 10, at which point further hits are not rewarded. This clearly helps it identify the critical context at which the fortress becomes vulnerable. 
    \item We give the agent a bonus reward of +2 for fortress destruction, to help it identify when the fortress is destroyed.
\end{itemize}

Table \ref{tab:easyresults} presents the results for agents trained after these 2 changes have been introduced to the reward structure to ease context identification. PPO with the recurrent architecture SF-GRU achieves the best performance in both score and number of fortress deaths, learning faster than SF-FF, and achieving a higher final score. The performance of both A2C and Rainbow also improves, although they are still unable to outperform humans. 
Fig. \ref{fig:ppo_curves} tracks the learning curves for PPO learning with all 3 reward settings (default, dense and after making context identification easier) in both game versions - Autoturn and Youturn. 


\subsection{Temporal Sensitivity}
Having established that context insensitivity is the primary driver of poor performance of state of the art RL algorithms on Space Fortress, we now analyze the temporal sensitivity of the PPO algorithm. As described in Section \ref{sec:gamedescription}, Space Fortress has a dominant temporal aspect - missiles must hit the fortress at least 250ms apart when it is not vulnerable, and then the strategy must reverse to hit the fortress \textit{twice within 250ms} when it is vulnerable, in order to destroy it. In order to understand whether the RL algorithms had developed any understanding of time as an independent dimension, we modified the critical time interval from 250ms to other values, and checked for positive transfer of learning from the policy trained with 250ms as the critical time interval. We achieve transfer of learning by simply initializing the weights of the transferee with the learned weights of the transferer.

Figure \ref{fig:temporal} compares the learning curves for an agent learning with PPO (using the SF-FF architecture) on Youturn, when the critical time interval is changed from 250ms to $\{125,400,600\}$ms. The blue line is for an agent learning from scratch, while the orange line is for an agent transferring learning from the PPO SF-FF agent trained on the 250ms interval. From Figure \ref{fig:temporal_400} and \ref{fig:temporal_600}, it can be seen that while the transfer of learning helps by initializing the weights in a favorable corner of the parameter space, the learning saturates very quickly and ends up with a final score much lower than achieved when the critical interval was 250ms. Modifying the critical time interval in Space Fortress is a useful technique to study temporal sensitivity of reinforcement learning algorithms.

\section{Conclusion}
This paper introduced Space Fortress as a new challenge for deep reinforcement learning research, with its time-sensitive game play, abrupt context-dependent shift of strategy and sparse rewards. 
We showed that state of the art RL algorithms (PPO, A2C and Rainbow) were unable to learn to play the game with neither the default sparse rewards nor the dense reward structure we defined. After making context identification easier through two minor tweaks in the reward structure, however, PPO was able to learn to play the game, outperforming humans comfortably. This ablation study allowed us to conclude that context insensitivity was the primary reason behind the poor performance of RL algorithms on Space Fortress, along with the inability to learn with sparse rewards. 
We then looked at whether PPO develops a concept of time as an independent variable - by checking for positive transfer of learning while changing the critical time interval of 250ms in Space Fortress. We found that while there was some positive transfer of learning, the agents saturated very quickly and did not achieve a good final score. By studying generalization and transfer across different settings of the critical time interval, Space Fortress can hence also be used as a benchmark to study temporal sensitivity of reinforcement learning algorithms.

Learning to play Space Fortress without making any modifications to the reward structure will require reinforcement learning algorithms to be able to identify various latent contexts and adapt their strategies suitably. It will also require being able to learn with very sparse rewards. This is beyond the capability of current state of the art reinforcement learning algorithms, making Space Fortress a useful benchmark for research.

\subsubsection*{Acknowledgments}
This research was sponsored by AFOSR Grant FA9550-15-1-0442. The collection of human data and development of the OpenAI Gym interface for Space Fortress was supported by ONR grant N00014-15-1-2151. We would like to thank Shawn Betts and John Anderson for insightful discussions on the game of Space Fortress, and for the OpenAI Gym interface for Space Fortress which was used to run the experiments in this paper.

\bibliography{refs}

\begin{thebibliography}{}

\bibitem[\protect\citeauthoryear{Anderson \bgroup et al\mbox.\egroup
  }{2018}]{anderson_betts_bothell_hope_lebiere_2018}
Anderson, J.; Betts, S.; Bothell, D.; Hope, R.~M.; and Lebiere, C.
\newblock 2018.
\newblock Three aspects of skill acquisition.

\bibitem[\protect\citeauthoryear{Anderson}{2009}]{anderson2009can}
Anderson, J.~R.
\newblock 2009.
\newblock {\em How can the human mind occur in the physical universe?}
\newblock Oxford University Press.

\bibitem[\protect\citeauthoryear{Andrychowicz \bgroup et al\mbox.\egroup
  }{2017}]{andrychowicz2017hindsight}
Andrychowicz, M.; Wolski, F.; Ray, A.; Schneider, J.; Fong, R.; Welinder, P.;
  McGrew, B.; Tobin, J.; Abbeel, O.~P.; and Zaremba, W.
\newblock 2017.
\newblock Hindsight experience replay.
\newblock In {\em Advances in Neural Information Processing Systems},
  5048--5058.

\bibitem[\protect\citeauthoryear{Bacon, Harb, and
  Precup}{2017}]{bacon2017option}
Bacon, P.-L.; Harb, J.; and Precup, D.
\newblock 2017.
\newblock The option-critic architecture.
\newblock In {\em AAAI},  1726--1734.

\bibitem[\protect\citeauthoryear{Bellemare \bgroup et al\mbox.\egroup
  }{2013}]{bellemare2013arcade}
Bellemare, M.~G.; Naddaf, Y.; Veness, J.; and Bowling, M.
\newblock 2013.
\newblock The arcade learning environment: An evaluation platform for general
  agents.
\newblock {\em Journal of Artificial Intelligence Research} 47:253--279.

\bibitem[\protect\citeauthoryear{Bellemare \bgroup et al\mbox.\egroup
  }{2016}]{bellemare2016unifying}
Bellemare, M.; Srinivasan, S.; Ostrovski, G.; Schaul, T.; Saxton, D.; and
  Munos, R.
\newblock 2016.
\newblock Unifying count-based exploration and intrinsic motivation.
\newblock In {\em Advances in Neural Information Processing Systems},
  1471--1479.

\bibitem[\protect\citeauthoryear{Bellemare, Dabney, and
  Munos}{2017}]{bellemare2017distributional}
Bellemare, M.~G.; Dabney, W.; and Munos, R.
\newblock 2017.
\newblock A distributional perspective on reinforcement learning.
\newblock {\em arXiv preprint arXiv:1707.06887}.

\bibitem[\protect\citeauthoryear{Brockman \bgroup et al\mbox.\egroup
  }{2016}]{brockman2016openai}
Brockman, G.; Cheung, V.; Pettersson, L.; Schneider, J.; Schulman, J.; Tang,
  J.; and Zaremba, W.
\newblock 2016.
\newblock Openai gym.
\newblock {\em arXiv preprint arXiv:1606.01540}.

\bibitem[\protect\citeauthoryear{Cho \bgroup et al\mbox.\egroup
  }{2014}]{cho2014learning}
Cho, K.; Van~Merri{\"e}nboer, B.; Gulcehre, C.; Bahdanau, D.; Bougares, F.;
  Schwenk, H.; and Bengio, Y.
\newblock 2014.
\newblock Learning phrase representations using rnn encoder-decoder for
  statistical machine translation.
\newblock {\em arXiv preprint arXiv:1406.1078}.

\bibitem[\protect\citeauthoryear{Destefano and
  Gray}{2016}]{destefano2016should}
Destefano, M., and Gray, W.~D.
\newblock 2016.
\newblock Where should researchers look for strategy discoveries during the
  acquisition of complex task performance? the case of space fortress.
\newblock In {\em Proceedings of the 38th Annual Conference of the Cognitive
  Science Society},  668--673.

\bibitem[\protect\citeauthoryear{Fortunato \bgroup et al\mbox.\egroup
  }{2017}]{fortunato2017noisy}
Fortunato, M.; Azar, M.~G.; Piot, B.; Menick, J.; Osband, I.; Graves, A.; Mnih,
  V.; Munos, R.; Hassabis, D.; Pietquin, O.; et~al.
\newblock 2017.
\newblock Noisy networks for exploration.
\newblock {\em arXiv preprint arXiv:1706.10295}.

\bibitem[\protect\citeauthoryear{Hessel \bgroup et al\mbox.\egroup
  }{2018}]{hessel2017rainbow}
Hessel, M.; Modayil, J.; van Hasselt, H.; Schaul, T.; Ostrovski, G.; Dabney,
  W.; Horgan, D.; Piot, B.; Azar, M.; and Silver, D.
\newblock 2018.
\newblock Rainbow: Combining improvements in deep reinforcement learning.
\newblock {\em AAAI Conference on Artificial Intelligence}.

\bibitem[\protect\citeauthoryear{Krizhevsky, Sutskever, and
  Hinton}{2012}]{krizhevsky2012imagenet}
Krizhevsky, A.; Sutskever, I.; and Hinton, G.~E.
\newblock 2012.
\newblock Imagenet classification with deep convolutional neural networks.
\newblock In {\em Advances in neural information processing systems},
  1097--1105.

\bibitem[\protect\citeauthoryear{Kulkarni \bgroup et al\mbox.\egroup
  }{2016}]{kulkarni2016hierarchical}
Kulkarni, T.~D.; Narasimhan, K.; Saeedi, A.; and Tenenbaum, J.
\newblock 2016.
\newblock Hierarchical deep reinforcement learning: Integrating temporal
  abstraction and intrinsic motivation.
\newblock In {\em Advances in neural information processing systems},
  3675--3683.

\bibitem[\protect\citeauthoryear{Man{\'e} and Donchin}{1989}]{mane1989space}
Man{\'e}, A., and Donchin, E.
\newblock 1989.
\newblock The space fortress game.
\newblock {\em Acta psychologica} 71(1-3):17--22.

\bibitem[\protect\citeauthoryear{Mnih \bgroup et al\mbox.\egroup
  }{2015}]{mnih2015human}
Mnih, V.; Kavukcuoglu, K.; Silver, D.; Rusu, A.~A.; Veness, J.; Bellemare,
  M.~G.; Graves, A.; Riedmiller, M.; Fidjeland, A.~K.; Ostrovski, G.; et~al.
\newblock 2015.
\newblock Human-level control through deep reinforcement learning.
\newblock {\em Nature} 518(7540):529.

\bibitem[\protect\citeauthoryear{Mnih \bgroup et al\mbox.\egroup
  }{2016}]{mnih2016asynchronous}
Mnih, V.; Badia, A.~P.; Mirza, M.; Graves, A.; Lillicrap, T.; Harley, T.;
  Silver, D.; and Kavukcuoglu, K.
\newblock 2016.
\newblock Asynchronous methods for deep reinforcement learning.
\newblock In {\em International Conference on Machine Learning},  1928--1937.

\bibitem[\protect\citeauthoryear{Ng, Harada, and Russell}{1999}]{ng1999policy}
Ng, A.~Y.; Harada, D.; and Russell, S.
\newblock 1999.
\newblock Policy invariance under reward transformations: Theory and
  application to reward shaping.
\newblock In {\em ICML}, volume~99,  278--287.

\bibitem[\protect\citeauthoryear{Osband \bgroup et al\mbox.\egroup
  }{2016}]{osband2016deep}
Osband, I.; Blundell, C.; Pritzel, A.; and Van~Roy, B.
\newblock 2016.
\newblock Deep exploration via bootstrapped dqn.
\newblock In {\em Advances in neural information processing systems},
  4026--4034.

\bibitem[\protect\citeauthoryear{Pathak \bgroup et al\mbox.\egroup
  }{2017}]{pathak2017curiosity}
Pathak, D.; Agrawal, P.; Efros, A.~A.; and Darrell, T.
\newblock 2017.
\newblock Curiosity-driven exploration by self-supervised prediction.
\newblock In {\em International Conference on Machine Learning (ICML)}, volume
  2017.

\bibitem[\protect\citeauthoryear{Plappert \bgroup et al\mbox.\egroup
  }{2017}]{plappert2017parameter}
Plappert, M.; Houthooft, R.; Dhariwal, P.; Sidor, S.; Chen, R.~Y.; Chen, X.;
  Asfour, T.; Abbeel, P.; and Andrychowicz, M.
\newblock 2017.
\newblock Parameter space noise for exploration.
\newblock {\em arXiv preprint arXiv:1706.01905}.

\bibitem[\protect\citeauthoryear{Russakovsky \bgroup et al\mbox.\egroup
  }{2015}]{russakovsky2015imagenet}
Russakovsky, O.; Deng, J.; Su, H.; Krause, J.; Satheesh, S.; Ma, S.; Huang, Z.;
  Karpathy, A.; Khosla, A.; Bernstein, M.; et~al.
\newblock 2015.
\newblock Imagenet large scale visual recognition challenge.
\newblock {\em International Journal of Computer Vision} 115(3):211--252.

\bibitem[\protect\citeauthoryear{Schaul \bgroup et al\mbox.\egroup
  }{2015}]{schaul2015prioritized}
Schaul, T.; Quan, J.; Antonoglou, I.; and Silver, D.
\newblock 2015.
\newblock Prioritized experience replay.
\newblock {\em arXiv preprint arXiv:1511.05952}.

\bibitem[\protect\citeauthoryear{Schulman \bgroup et al\mbox.\egroup
  }{2015}]{schulman2015high}
Schulman, J.; Moritz, P.; Levine, S.; Jordan, M.; and Abbeel, P.
\newblock 2015.
\newblock High-dimensional continuous control using generalized advantage
  estimation.
\newblock {\em arXiv preprint arXiv:1506.02438}.

\bibitem[\protect\citeauthoryear{Schulman \bgroup et al\mbox.\egroup
  }{2017}]{schulman2017proximal}
Schulman, J.; Wolski, F.; Dhariwal, P.; Radford, A.; and Klimov, O.
\newblock 2017.
\newblock Proximal policy optimization algorithms.
\newblock {\em arXiv preprint arXiv:1707.06347}.

\bibitem[\protect\citeauthoryear{Silver \bgroup et al\mbox.\egroup
  }{2016}]{silver2016mastering}
Silver, D.; Huang, A.; Maddison, C.~J.; Guez, A.; Sifre, L.; Van Den~Driessche,
  G.; Schrittwieser, J.; Antonoglou, I.; Panneershelvam, V.; Lanctot, M.;
  et~al.
\newblock 2016.
\newblock Mastering the game of go with deep neural networks and tree search.
\newblock {\em nature} 529(7587):484--489.

\bibitem[\protect\citeauthoryear{Sutskever, Vinyals, and
  Le}{2014}]{sutskever2014sequence}
Sutskever, I.; Vinyals, O.; and Le, Q.~V.
\newblock 2014.
\newblock Sequence to sequence learning with neural networks.
\newblock In {\em Advances in neural information processing systems},
  3104--3112.

\bibitem[\protect\citeauthoryear{Todorov, Erez, and
  Tassa}{2012}]{todorov2012mujoco}
Todorov, E.; Erez, T.; and Tassa, Y.
\newblock 2012.
\newblock Mujoco: A physics engine for model-based control.
\newblock In {\em Intelligent Robots and Systems (IROS), 2012 IEEE/RSJ
  International Conference on},  5026--5033.
\newblock IEEE.

\bibitem[\protect\citeauthoryear{Towne, Boot, and
  Ericsson}{2016}]{towne2016understanding}
Towne, T.~J.; Boot, W.~R.; and Ericsson, K.~A.
\newblock 2016.
\newblock Understanding the structure of skill through a detailed analysis of
  individuals' performance on the space fortress game.
\newblock {\em Acta psychologica} 169:27--37.

\bibitem[\protect\citeauthoryear{van~der Linden \bgroup et al\mbox.\egroup
  }{2017}]{van2017higher}
van~der Linden, P.~A.; Intelligentie, B. O.~K.; Poppinga, G.; Roessingh, J.;
  and van Splunter, S.
\newblock 2017.
\newblock On higher-order control tasks: The application of a3c on space
  fortress.

\bibitem[\protect\citeauthoryear{van Oijen \bgroup et al\mbox.\egroup
  }{2017}]{van2017towards}
van Oijen, J.; Poppinga, G.; Brouwer, O.; Aliko, A.; and Roessingh, J.~J.
\newblock 2017.
\newblock Towards modeling the learning process of aviators using deep
  reinforcement learning.
\newblock In {\em Systems, Man, and Cybernetics (SMC), 2017 IEEE International
  Conference on},  3439--3444.
\newblock IEEE.

\bibitem[\protect\citeauthoryear{Wang \bgroup et al\mbox.\egroup
  }{2015}]{wang2015dueling}
Wang, Z.; Schaul, T.; Hessel, M.; Van~Hasselt, H.; Lanctot, M.; and De~Freitas,
  N.
\newblock 2015.
\newblock Dueling network architectures for deep reinforcement learning.
\newblock {\em arXiv preprint arXiv:1511.06581}.

\bibitem[\protect\citeauthoryear{Zambrano, Roelfsema, and
  Bohte}{2015}]{zambrano2015continuous}
Zambrano, D.; Roelfsema, P.~R.; and Bohte, S.~M.
\newblock 2015.
\newblock Continuous-time on-policy neural reinforcement learning of working
  memory tasks.
\newblock In {\em Neural Networks (IJCNN), 2015 International Joint Conference
  on},  1--8.
\newblock IEEE.

\end{thebibliography}
\bibliographystyle{aaai}

\end{document}